\documentclass{ieeetj}
\usepackage{cite}
\usepackage{amsmath,amssymb,amsfonts}
\usepackage{algorithmic}
\usepackage{graphicx,color}
\usepackage{textcomp}
\usepackage{xcolor}
\usepackage{hyperref}
\usepackage{algorithm,algorithmic}

\usepackage{subcaption}
\usepackage{adjustbox}
\usepackage{wrapfig}
\usepackage{soul}

\usepackage{booktabs}

\usepackage[capitalize]{cleveref}

\usepackage[capitalize]{cleveref}
\crefname{section}{Sec.}{Secs.}
\Crefname{section}{Section}{Sections}
\Crefname{table}{Table}{Tables}
\crefname{table}{Tab.}{Tabs.}

\newcommand{\norm}[1]{\left \lVert #1 \right \rVert}

\def\BibTeX{{\rm B\kern-.05em{\sc i\kern-.025em b}\kern-.08em
    T\kern-.1667em\lower.7ex\hbox{E}\kern-.125emX}}
\AtBeginDocument{\definecolor{tmlcncolor}{cmyk}{0.93,0.59,0.15,0.02}\definecolor{NavyBlue}{RGB}{0,86,125}}

\def\authorrefmark#1{\ensuremath{^{\textbf{#1}}}}

\begin{document}

\markboth{}{Bar and Giryes}

\title{Diverse Subset Selection via Norm-Based Sampling and Orthogonality }

\author{Noga Bar\authorrefmark{1} and Raja Giryes \authorrefmark{1}}
\affil{School of Electrical Engineering at Tel Aviv University}
\corresp{Corresponding author: nogabar@mail.tau.ac.il.}
\authornote{This work was supported by The Center for AI \& Data Science at Tel Aviv University (TAD).}

\begin{abstract}
Large annotated datasets are crucial for the success of deep neural networks, but labeling data can be prohibitively expensive in domains such as medical imaging. This work tackles the \textit{subset selection problem}: selecting a small set of the most informative examples from a large unlabeled pool for annotation.
We propose a simple and effective method that combines feature norms, randomization, and orthogonality (via the Gram–Schmidt process) to select diverse and informative samples. Feature norms serve as a proxy for informativeness, while randomization and orthogonalization reduce redundancy and encourage coverage of the feature space.
Extensive experiments on image and text benchmarks, including CIFAR-10/100, Tiny ImageNet, ImageNet, OrganAMNIST, and Yelp, show that our method consistently improves subset selection performance, both as a standalone approach and when integrated with existing techniques.
\end{abstract}



\maketitle

\section{INTRODUCTION}

Deep neural networks owe much of their success to the availability of large, annotated datasets. However, acquiring labeled data can be challenging especially in specialized domains like medical imaging, where expert clinicians must perform the annotation at considerable expense.
Consequently, when resources are limited, it becomes crucial to select the most informative samples to be annotated from a large unlabeled pool. This challenge is referred to as the \emph{subset selection problem}.

The subset selection problem is closely related to active learning, which involves gradually selecting unlabeled examples for annotation during the learning process.
Unlike active learning, unsupervised subset selection focuses on scenarios in which the entire set of samples to be annotated must be decided all at once.
This single-shot approach poses several challenges. Specifically, identifying informative samples requires balancing criteria such as diversity, relevance, and coverage of the data distribution, which is far from trivial.
Techniques based on simple heuristics such as uncertainty, entropy, or margin between the highest scores have been demonstrated to be ineffective for extremely small subsets.
Moreover, many methods proposed for subset selection have failed to outperform random sampling, particularly when a very small set of examples is chosen to be annotated \cite{hacohen2022active,chen2022making,guo2022deepcore}.

\noindent\textbf{Norms and Randomness for Selection.} The task of selecting extremely small labeled subsets presents a fundamental challenge: choosing examples that are both representative and diverse, in order to capture the underlying data distribution effectively.
In this work, we propose a simple yet effective approach based on two core principles: feature norms and randomness.
We assume access to an informative and unsupervised feature extractor, where a high feature norm indicates strong alignment with the learned representation space. Thus, feature norms serve as a proxy for informativeness, while randomness is introduced to promote diversity among the selected samples.

These approaches are closely relate to pruning neural networks, where a small set of neurons are selected according to their norms \cite{he2018soft, he2019filter,li2017pruning}. Furthermore, is was shown that using randomization in the pruning process enhances performance \cite{bar2023pruning} and is often better than deterministic pruning.

\noindent\textbf{Gram–Schmidt.} While feature norms provide a useful signal for informativeness, relying on them alone may lead to selecting redundant, highly correlated examples. To mitigate this, we incorporate the Gram–Schmidt process to iteratively select examples that are orthogonal to those already chosen. Specifically, at each step, we target high-norm examples in the residual subspace—that is, the space orthogonal to the span of previously selected features. This procedure explicitly accounts for correlations among data points and encourages broader coverage of the feature space.

\noindent\textbf{Motivational Example.}
We illustrate the motivation behind our method using the classic EigenFaces algorithm \cite{sirovich1987low, turk1991face}, which performs face recognition by projecting images into a linear feature space using PCA.
The results highlight the importance of combining multiple components: without randomization, the selected examples lack diversity and fail to cover the feature space; without norm-based selection, the chosen examples are less informative; and with the addition of the Gram–Schmidt process, the selection achieves broader coverage by reducing redundancy.

\noindent\textbf{Empirical Results.} Extending the linear insights from linear and small scale case, to non-linear embeddings, we aim to preserve high norms within feature spaces induced by neural networks.
Specifically, we apply our approach to features generated by fully random networks (at initialization, without training) and self-supervised networks, namely, SimCLR \cite{chen2020simple} and DINO \cite{caron2021emerging},
on the CIFAR-10/100, Tiny-ImageNet, ImageNet, and OrganAMNIST \cite{medmnistv2} datasets.
We also evaluate our approaches with language task using the Yelp dataset with BERT \cite{devlin2018bert}.
Our experiments confirm that our norm- and orthogonality-driven approach yields robust subset selection across both diverse domains.

Notably, integrating our norm-based approach with existing subset selection methods often leads to further performance gains, achieving state-of-the-art results in many settings. Our method is versatile and consistently effective across different training frameworks and feature domains. Finally, a comprehensive ablation study highlights the robustness of the norm criterion, reinforcing its value as a simple and reliable tool for subset selection.

\section{RELATED WORK}\label{sec:related}
\textbf{Subset Selection vs. Active Learning.} Our work focuses on the \emph{unsupervised subset selection problem}. While both subset selection and active learning aim to identify informative examples for labeling, they differ in methodology. Subset selection involves choosing a fixed subset of examples from the entire dataset, whereas active learning dynamically selects examples across multiple iterations based on the current model.  
Unsupervised subset selection is particularly useful when labeling is expensive and must be performed in a single batch. In contrast, active learning requires multiple interactions with an annotator throughout the training process. Importantly, subsets in the subset selection setting should be model-agnostic and enable strong performance when training a model from scratch. Although some active learning methods can be adapted for single-iteration use without relying on previously labeled data, they are not inherently designed for the subset selection setting. 

\noindent\textbf{Common Approaches for Selection.}
A variety of strategies have been explored for selecting representative and diverse examples. Common approaches include uncertainty sampling \cite{gal2017deep,settles2009active}, diversity maximization \cite{ash2019deep,citovsky2021batch,beluch2018power,hsu2015active,chang2017active,chen2020simple}, and entropy-based criteria \cite{coleman2019selection}. These methods generally aim to capture the underlying data distribution while minimizing redundancy among the chosen examples. 

In active learning, many methods are tailored to the geometry of the data representation. Popular approaches rely on clustering (often k-means) to select diverse sets \cite{hacohen2022active,sener2018active,sorscher2022beyond,xia2022moderate} or aim to ensure coverage of the embedding space \cite{yehuda2022active,zheng2022coverage}.  
Optimization-based formulations have also been proposed, where the goal is to approximate the gradient of the full dataset with a small, informative subset \cite{borsos2020coresets,paul2021deep,killamsetty2021glister,killamsetty2021grad,mirzasoleiman2020coresets,yang2023towards}. Another line of work combines labeled datasets with pretrained representations and curriculum learning to design model-agnostic subset selection methods \cite{killamsetty2023milo}.  

All of these methods focus on relatively large labeled subsets (at least 10\% of the data). In contrast, our focus is on selection of \emph{extremely small subsets} from \emph{unsupervised dataset}, where the number of selected examples to label is on the same order as the number of classes. 

\noindent\textbf{Challenges and Benchmarks. } Recent challenges in subset selection focus on choosing examples from a labeled pool rather than an unlabeled one \cite{gadre2024datacomp,mazumder2024dataperf}. Moreover, a prior work introduced a framework selection \cite{guo2022deepcore}.
 
\noindent\textbf{Low-Budget Subset Selection.}
Closest to our work are ProbCover \cite{yehuda2022active} and TypiClust \cite{hacohen2022active}, which represent state-of-the-art approaches for extremely small subset selection in image classification.  
ProbCover leverages advances in self-supervised learning to enrich data representations and maximize probability coverage, thereby identifying examples that contribute most information under low-budget annotation. TypiClust, on the other hand, exploits a phase-transition phenomenon: in low-budget regimes, typical examples provide the most benefit, while in larger budgets, atypical examples are more useful.  

Although effective, both methods depend heavily on feature quality and require costly computations such as clustering or adjacency graph construction. In contrast, our approach is less domain-dependent and computationally lighter, relying instead on simple yet effective concepts such as feature norms, feature orthogonality, and randomization to balance informativeness and diversity.

\section{SUBSET SELECTION}
\noindent\textbf{Problem Statement. }
Given an unlabeled large dataset with $D = \{x_1, ..., x_N\}$ the goal is to select a small subset of $s \ll N$  examples. The selected examples form a labeled small set,  $S = \{(x_1, y_1),...(x_s, y_s)\}$.
The objective is to maximize the performance of the model after training, utilizing these small labeled subset.
We assume that each unlabeled example, $x_i$, has a corresponding feature vector, $F_i$, which is derived from a feature extractor, $\mathcal{M}$.
This extractor is typically a model trained in an unsupervised manner or pre-trained on a different dataset, such that $F_i = \mathcal{M}(x_i;\theta)$, learned using $D$ or another unlabeled data.

In our setting, we do not assume access to the model $\mathcal{M}$ itself, only to its output features and the raw unlabeled data. This constraint necessitates making labeling decisions based solely on feature representations. Although strong performance is generally difficult to achieve with such a small labeled set without leveraging advanced semi- or unsupervised learning techniques, this setup offers significant efficiency benefits.
In particular, training on the small subset can converge rapidly. We show that even under these limitations, simple tools can yield meaningful performance gains.
Hence, the proposed setting aligns naturally with practical scenarios in which obtaining labels is difficult and resource efficiency is a key concern.

Other closely related problems share the goal of reducing the data load during training but differ in their settings and motivations. \textit{Active learning} incrementally expands the labeled set by querying informative examples—typically based on uncertainty or diversity—after supervised training has begun \cite{settles2009active, sener2018active}
In addition, \textit{dataset pruning} aims to reduce the training set size by removing redundant or uninformative examples, frequently borrowing techniques from neural network pruning. However, these methods usually assume full access to labels and retain large portions of the original data, often at least 20\% \cite{yang2022dataset}. In contrast, the \textit{data attribution} task aims to characterize the examples and provide explainability to models after training.
Finally, \textit{curriculum learning and data cartography} aim to order training examples based on their importance or difficulty, facilitating optimization rather than directly reducing data volume \cite{bengio2009curriculum,swayamdipta2020dataset}.

\noindent \textbf{Feature Norm.} 
The first selection strategy we explore is based on the norms of neural features.
Let $F_1, ..., F_N$ denote the features corresponding to $N$ unlabeled training examples.
These features are obtained from a neural network and refer specifically to the output of the penultimate layer (i.e., just before the final linear classification layer).
We focus on the norms of these features, rather than on properties of the raw input. To select examples, we sample them at random according to a probability distribution proportional to the feature norms:
\begin{align}\label{eq:prob_norm}
    p_i = \frac{\norm{F_i}}{\sum_{j=1}^N \norm{F_j}}, \quad i=1,...,N,
\end{align}
where $\norm{\cdot}$ is the $\ell_2$ norm unless otherwise stated.
In \cref{sec:exp}, we show that this simple strategy outperforms uniform random selection—a strong non-trivial baseline in the regime of extremely small labeled subsets \cite{hacohen2022active, chen2022making}. Moreover, we show that there is an advantage to randomization over deterministic selection of the highest norms.

\noindent\textbf{Motivation for neural features norm for selection.}
Given a meaningful feature extractor $\mathcal{M}(x;\theta)$, the norm of the extracted feature for each input $F_i = \mathcal{M}(x_i;\theta)$ can serve as an indicator of how well an input $x_i$ aligns with the feature space defined by $\mathcal{M}$.
A higher norm suggests that $x_i$ exhibits stronger correlation and alignment with the representation. When the feature space is  informative and aligned with the data, larger norms often correlate with more salient examples.

\noindent\textbf{Subset Selection as Pruning.}
Another perspective motivating the use of feature norms comes from drawing an analogy between input subset selection and network pruning. Selecting input examples is similar to selecting filters, with unselected examples treated as zeroed-out inputs that do not participate in training.
Pruning literature often emphasizes the importance of high-norm components: high-norm filters in structured pruning \cite{he2018soft, he2019filter}, high-magnitude weights in unstructured pruning \cite{frankle2018lottery}, and high-norm features for layer-wise pruning \cite{li2017pruning}. Inspired by these findings, we adopt a norm-based criterion for input selection.

\noindent \textbf{Randomization.}
Instead of deterministically selecting the examples with the highest feature norms, we incorporate randomization into the selection process by sampling according to the distribution in \cref{eq:prob_norm}. This probabilistic approach avoids redundancy and promotes diversity among selected samples.
Randomization has been shown to be beneficial in related settings such as neural pruning, where it helps improve performance by avoiding overly greedy decisions \cite{bar2023pruning}.
We argue that using randomization is crucial for achieving good performance.

\begin{algorithm}[t]

\captionof{algorithm}{Gram–Schmidt for Subset Selection}\label{algorithm}
   \hrule
\begin{algorithmic}[width=\columnwidth]
   \STATE \textbf{Input:} 
   Feature extractor model $\mathcal{M}$, unlabeled examples $\{x_i\}_{i=1}^N$
   and $s$ number of examples to label. 
   \STATE \textbf{Feature Extraction:} $F_i = \mathcal{M}(x_i)$
   \STATE {\bfseries Initialize:} $\Tilde{F}_i = F_i$ and $S=\emptyset$
   \FOR{$k=1$ {\bfseries to} $s$} 
   \STATE \textbf{Randomize} $i$ according to $p_j = \frac{\norm{\Tilde{F}_j}} {\sum_{k=1}^N \norm{\Tilde{F}_k}}$, $j\notin S$
   \STATE \textbf{Update:} $S = S \cup \{i\}$
   \STATE \textbf{Projection:} For $j\notin S$: $\Tilde{F}_j = \Tilde{F}_j - \frac{\Tilde{F}_j^T \Tilde{F}_i}{\norm{\Tilde{F}_i}^2} \Tilde{F}_i $
   \ENDFOR
   \RETURN $S$
\end{algorithmic}
\end{algorithm}

\noindent \textbf{Gram-Schmidt.} In the context of subset selection, it is crucial to emphasize that solely relying on feature norms may not yield optimal results.
While norms provide valuable information, they may not capture the full complexity of the dataset.
Specifically, the norm values do not contain information about the correlations between data points.
Therefore, it is essential to augment norm-based selection methods with additional concepts to enhance their effectiveness.
We leverage techniques from linear algebra and select features that span the domain. Namely, we utilize the Gram-Schmidt process described in \cref{algorithm} to iteratively choose orthogonal features. 
The first step of the selection is to randomly draw an example to be labeled according to the norms using the probabilities in \cref{eq:prob_norm}.
Then, we update the set of chosen examples, $S$.
Finally, in the Gram–Schmidt step, we remove the projection of the selected feature from the remaining (unselected) features, ensuring that the residuals are orthogonal to the chosen directions. We repeat the selection and projection steps iteratively until reaching the desired subset size. This process explicitly accounts for correlations among data points.
Thus, we ensure that the chosen examples have not only high feature norms but also capture diverse and informative aspects of the dataset.

\noindent \textbf{Complexity.} Our methods enjoy low computational complexity.
Let $d$ be the dimension of the features. First, we calculate the normalized vector of probabilities which takes $O(Nd)$.
Then, in each selection iteration we sample according to the norms (\cref{eq:prob_norm}), the complexity of weighted sampling of a vector of dimension $N$ is $O(\log N)$ and the sampling is performed $s$ times.
Hence, the overall complexity of norm randomization is $O(Nd + s\log N)$.
For the Gram-Schmidt algorithm (\cref{algorithm}), the complexity of the projection step is $O(Nd)$ for calculating the inner products of the remaining features with the chosen feature. The calculation of the probability density vector and sampling takes $O(Nd+\log(N))=O(Nd)$. Overall, the complexity of choosing $s$ examples is $O(sNd)$.
In comparison, the baseline, ProbCover \cite{yehuda2022active}, requires $O(N^2d)$ computations. 
In our complexity analysis, we assume that the features are given and we do not include their query to our computations since we rely on their availability (as done in previous work \cite{yehuda2022active}).

\subsection{Motivational Example}\label{sec:warmup}
In this section, we demonstrate the benefits of our methods using the simple EigenFaces algorithm \cite{sirovich1987low,turk1991face}. This demonstration will provide also an extra motivation for our proposed approach. 
EigenFaces is a classical algorithm used for face recognition.
In this problem, we have $x_1,.., x_N$ images of faces from $p$ different people and the person in each image is labeled.
The labels are denoted by $y_1, ..., y_N$.
The pre-processing step is simply calculating the PCA of the input: $W = \text{PCA}(x_1,..,x_N)$ and projecting the input to the main principal components, which hold the important information of the data. 
Thus, the PCA is used as an informative feature extractor. The projection of the input using $W$ results in the features: $\{Wx_i\}_{i=1}^N$, where we abuse notation here and denote by $W$ the projection onto the main components of the PCA. 
For classification, the model returns simply the class of the nearest neighbor after the PCA projection. Formally,
\begin{align}\label{eq:eigenface}
    j = \text{arg}\min_{i\in[N]} \{\norm{Wx_i - Wx}\}, \hspace{0.5cm} \text{Pred}(x) = y_j.
\end{align}

For subset selection, $W$ is the unsupervised feature extractor and 
we choose the subset according to the features, $\{Wx_i\}_{i=1}^N$, to form a small subset $S$. We calculate a new linear transformation, $W_s$, based only on the selected examples. 
Following the selection step, the labels of the chosen small subset are utilized to classify the faces (\cref{eq:eigenface}).

For this problem, selecting examples with high energy values (i.e., higher norms) after PCA projection is a natural choice as they are better aligned with the feature domain. 
Because PCA provides a simple yet informative representation of the feature space, high-energy examples tend to correspond to more informative data. Consequently, selecting a small set of  high-energy examples for classification effectively minimizes performance degradation associated with using small subsets.

It is important to note that relying solely on the norm criterion may result in selecting examples predominantly from a single class, particularly if that class has higher energy than others. To address this, ensuring diversity in the selection is crucial. This can be achieved by incorporating randomization into the selection process or by decorrelating the information already present in the selected examples, as is done with the Gram-Schmidt approach.

We asses our strategies with the EigenFaces algorithm and AT\&T dataset of faces. This dataset contains 400 images of faces of 40 people. We randomly split the data into 240 train and 160 test balanced sets.
We apply our approaches of randomization according to norm (\cref{eq:prob_norm}) and Gram-Schmidt (\cref{algorithm}). We compare them with uniform random sampling and selection according to the maximal norm.
We report accuracy and class coverage.

\begin{table}[t]
\centering
\begin{adjustbox}{width=1\linewidth}
\begin{tabular}{lcc|cc}
\toprule
Method                & Accuracy & Coverage & Accuracy & Coverage \\ 
Subset size & \multicolumn{2}{c|}{40} & \multicolumn{2}{c}{80} \\ \midrule
Random          & 43.13             & 27                & 65.00             & 35                \\ 
Max Norm        & 30.00             & 16                & 45.63             & 23                \\ 
Norm Randomization & 49.38          & 26                & 71.25             & \textbf{37 }               \\ 
Gram-Schmidt    & \textbf{58.13}             & \textbf{32}                & \textbf{75.00}             & \textbf{37 }               \\ \bottomrule
\end{tabular}
\end{adjustbox}

\caption{\textbf{Warmup example:} Face recognition with EigenFaces. Comparison of accuracy and class coverage. The number of classes is $40$.}
\label{tab:toy_example}
\vspace{-0.2in}
\end{table}

The results in \cref{tab:toy_example} show the benefits of our methods. 
Norm randomization induces good results and coverage. Norm without randomization suffers from a major degradation in coverage which is probably the cause of its failure.
The accuracy and the coverage are highest with Gram-Schmidt supporting the hypothesis that a good representation of the domain is beneficial.
Additionally, it is clear that the performance does not only rely on the coverage of the classes, but also the quality of the chosen examples is crucial. This can be observed in the case of selecting 40 examples where random sampling has better coverage but norm randomization achieves better accuracy. Also, for 80 examples Gram-Schmidt enjoys better performance than norm randomization even with the same class coverage.
Having motivated our proposed framework in the linear case, we move test it in the non-linear case.

\begin{figure*}[t]
  \centering
  \includegraphics[width=1.\textwidth]{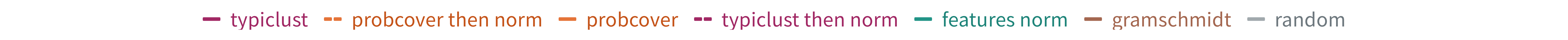}
  \begin{subfigure}{0.24\linewidth}
    \includegraphics[width=1\textwidth]{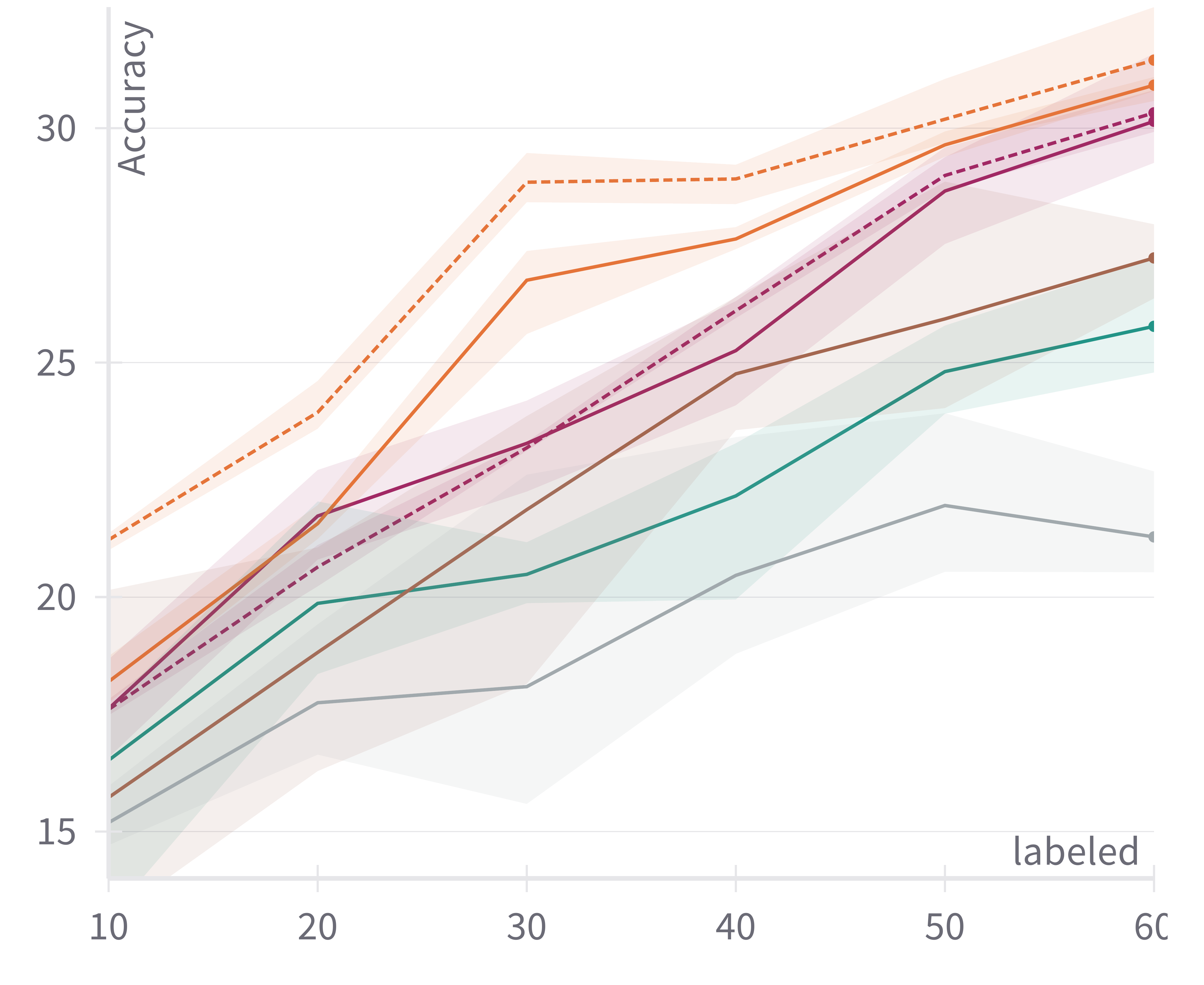}
    \caption{CIFAR-10}
    \label{fig:fs_cifar10}
  \end{subfigure}
  \begin{subfigure}{0.24\linewidth}
    \includegraphics[width=1\textwidth]{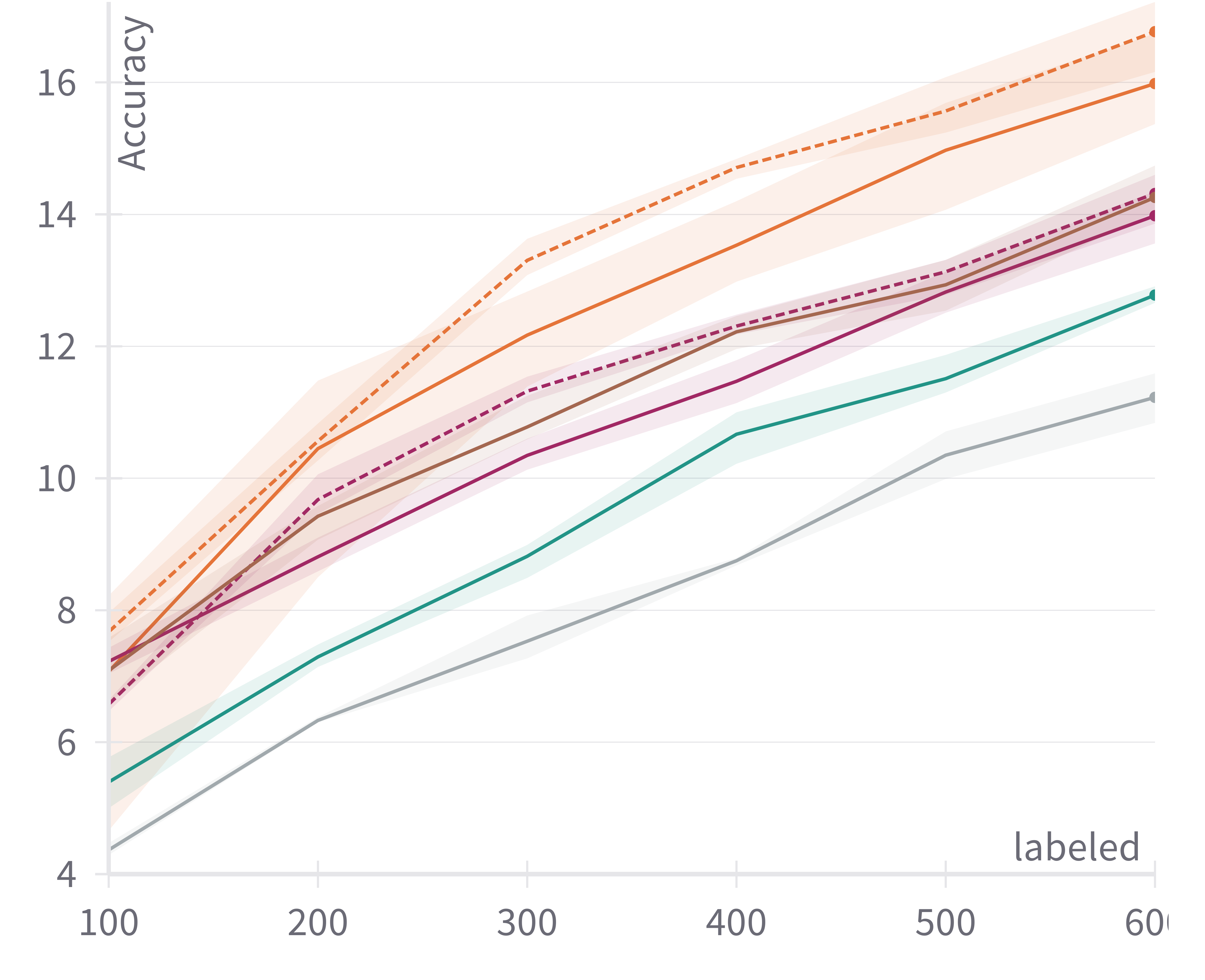}
    \caption{CIFAR-100}
    \label{fig:fs_cifar100}
  \end{subfigure} 
    \begin{subfigure}{0.24\linewidth}
    \includegraphics[width=1\textwidth]{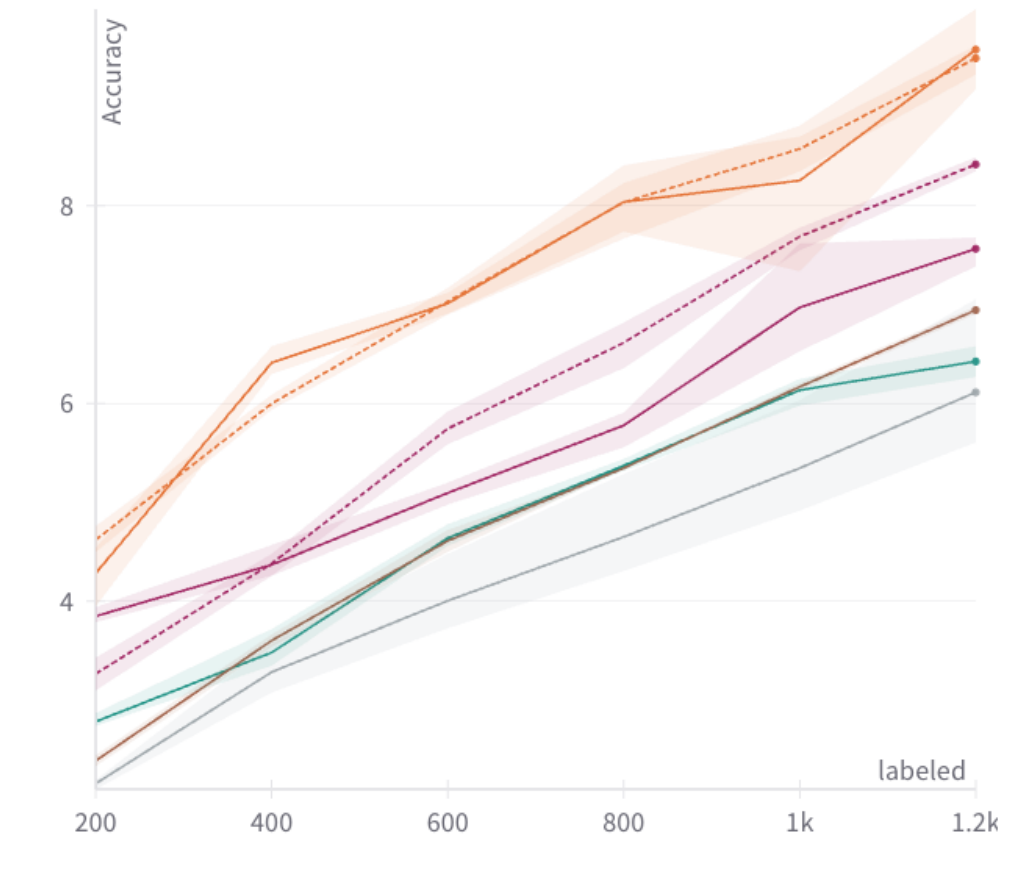}
    \caption{Tiny-ImageNet}
    \label{fig:fs_tinyimagenet}
  \end{subfigure}
  \begin{subfigure}{0.24\linewidth}
    \includegraphics[width=1\textwidth]{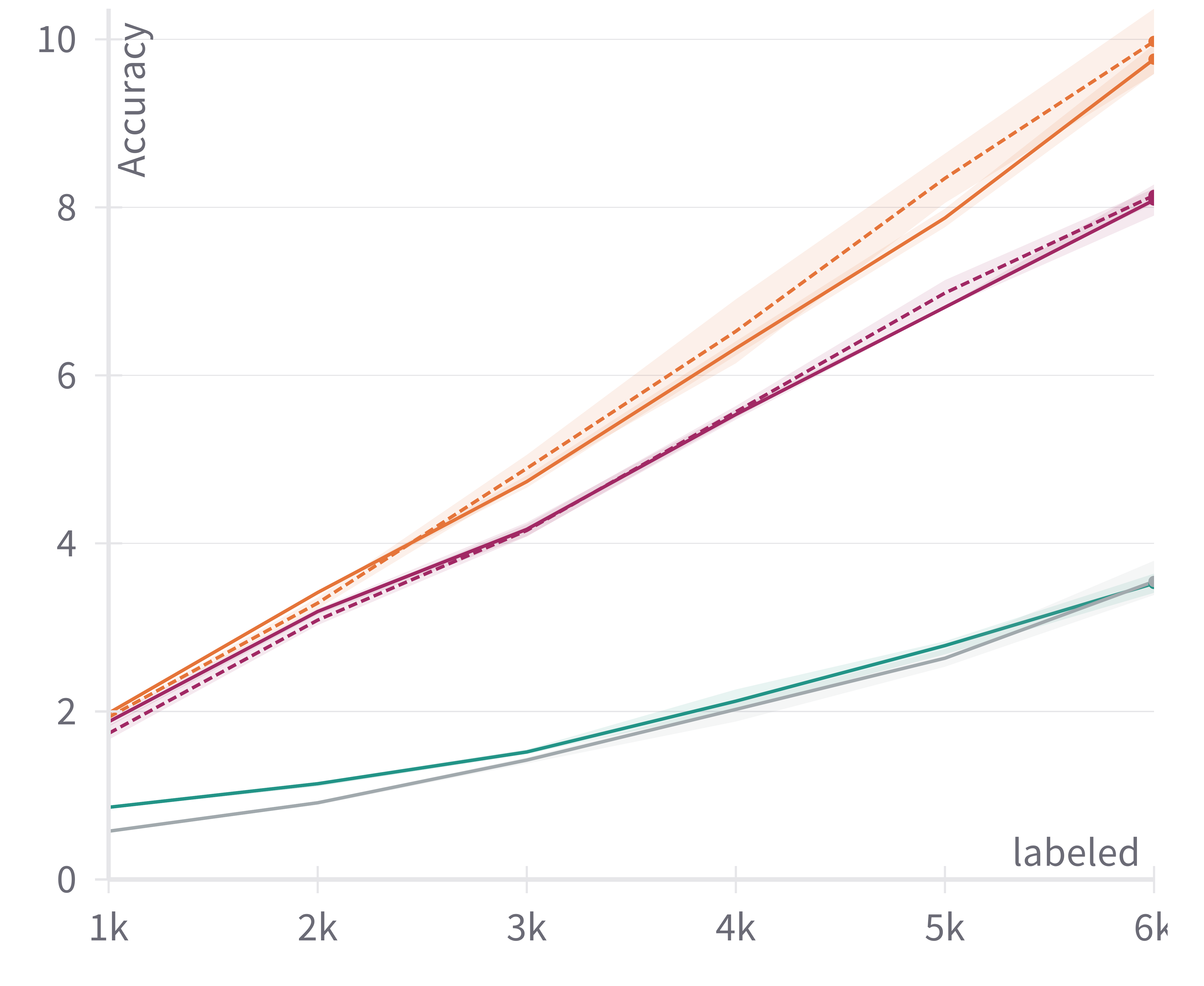}
    \caption{ImageNet}
    \label{fig:fs_imagenet}
  \end{subfigure}
  \vspace{-0.1in}
  \caption{\textbf{Fully-supervised framework:} The performance comparison includes both of our methods, randomization with feature norms and Gram-Schmidt, random selection, the baselines TypiClust and ProbCover, and the addition of our norm criterion to these baselines. An average of 3 seeds is presented and the shaded areas correspond to the standard error.}
  \label{fig:fs}
   \vspace{-0.5cm}
\end{figure*}

\section{EXPERIMENTS}
\label{sec:exp}
We validated our proposed strategy empirically across diverse settings and datasets, focusing on extremely small subsets selected from unlabeled pools. Consequently, in many instances, not all classes are represented in the labeled training set, leading to potential class imbalances of the subsets.

We evaluate our method in three frameworks: (i) \emph{Fully supervised}: Training exclusively with annotated data using an initialized ResNet-18 from scratch.
(ii) \emph{Semi-supervised with linear probing}: Training a single-layer linear classifier on top of self-supervised features obtained from an unlabeled dataset. This framework aims to utilize semi-supervised learning principles and utilize the informative feature space without relying on advances in pseudo-labeling techniques.
(iii) \emph{Semi-supervised}: Training competitive semi-supervised methods using subsets chosen by the selection algorithms. We employ FlexMatch \cite{zhang2021flexmatch} and  SimMatch \cite{zheng2022simmatch} to assess the effectiveness of our selection algorithm.

We present results with randomization according to the feature norms (as in \cref{eq:prob_norm}), the Gram-Schmidt based strategy (\cref{algorithm}), and a combination of ProbCover \cite{yehuda2022active} and TypiClust \cite{hacohen2022active} with the norm criterion.
ProbCover and TypiClust rely on self-supervised embeddings.
ProbCover is a method which selects examples that contribute the most information to the learning process and TypiClust prioritizes typical examples to improve the performance with extremely low budgets.
We combined the methods with the norm criterion and GS (\cref{algorithm}), named ``\textit{$<$method$>$ then norm}'' or ``\textit{$<$method$>$ then GS}'', by selecting twice the required budget, denoted as $2b$, and then randomly selecting $b$ examples based on feature norms.
We choose to embed the norm criterion this way since the methods rely on normalized features and are very sensitive to changes in the feature domain.

The same methodology of combining norms and Gram-Schmidt can be used for other existing methods when features are available.
Selecting examples based on their features introduces only a minor computational overhead to these methods. As we demonstrate below, this added computation leads to performance improvement in the vast majority of cases.
In the paper, we only show the combination of the feature norm criterion with these methods. 
When combined with the baseline, using Gram-Schmidt enhances the performance mainly in setting (ii) with a linear classifier.

For frameworks (i) and (ii), we use code adapted from \cite{yehuda2022active,hacohen2022active}, which is based on prior works \cite{vangansbeke2020scan,Munjal2020TowardsRA}.
For the semi-supervised setting (iii), we employ code of the Semi-Supervised benchmark \cite{usb2022}.
Our code is attached to the paper.

We compare our method with other subset selection methods: Uniform at random, which is a competitive baseline, TypiClust \cite{hacohen2022active} and ProbCover \cite{yehuda2022active}.
We do not include other subset selection methods since many of them do not surpass random selection and others are only slightly better than random selection, as demonstrated in \cite{guo2022deepcore,hacohen2022active,chen2022making}. Other methods are tailored for larger subsets while we focus on extremely small subsets.

We test our method on CIFAR-10, CIFAR-100, Tiny-ImageNet, ImageNet, Yelp and OrganAMNIST \cite{medmnistv2} datasets.
Specifically, unless otherwise specified, we utilized SimCLR \cite{chen2020simple} embeddings for CIFAR-10/100, Tiny-ImageNet and OrganAMNIST. 
For ImageNet, we use DINO \cite{caron2021emerging} embeddings.
We use the SimCLR implementation from \cite{vangansbeke2020scan} and train ResNet-18 with an MLP projection layer for $500$ epochs.
Post-training, we extract the 512-dimensional features from the penultimate layer.
We train the models with SGD with momentum and an initial learning rate of 0.4 with a cosine scheduler.
We employ a batch size of 512 and a weight decay of $10^{-4}$.
We augment the data with random resize and crop, random horizontal flips, color jittering, and random gray-scale.
For DINO, we use ViT-S/16 model pre-trained on ImageNet. 

In frameworks (i) and (ii), we evaluated the methods with varying numbers of examples: $[b, 2b, ..., 6b]$, where $b$ represents the number of classes. We report the average results over 3 seeds.
We focus on extremely small subsets and we compare our method with baselines that perform well with these sizes \cite{hacohen2022active,yehuda2022active}.
For the semi-supervised framework, (iii), we use $b$ labeled examples.

In the fully-supervised framework for CIFAR-10/100, Tiny-ImageNet, and OrganAMNIST, we train the model for 200 epochs using the SGD optimizer with Nesterov momentum and a cosine learning rate with an initial step-size of 0.025. 
We utilize a batch size of $\max\{\#\text{labeled}, 100\}$ and a weight decay of $3\times10^{-4}$. Augmentations include random crop and random horizontal flip.
For ImageNet, we use the same hyper-parameters as described above, with the exception that we train for 100 epochs and employ a batch size of 50 due to computational constraints.
We choose the best epoch according to a validation test.
For experimenting with Yelp, we fine-tune a pre-trained BERT \cite{devlin2018bert}.
We train for 100 epochs with AdamW optimizer and initial learning rate of $5\times10^{-5}$.
Instead of selecting the epoch with the highest validation accuracy, we evaluate the test set accuracy based on the last epoch. This choice reduces the computational overhead of repeated calculations of validation accuracy for a large dataset. Additionally, due to limited GPU memory we do not include results with ProbCover.
For the unsupervised features we use the embedding of the mid token of each example. This approach avoids the extremely high dimensionality that would result from unfolding the embeddings of all tokens.
We employed a single NVIDIA RTX-2080 GPU to undertake the learning processes, encompassing both the acquisition of self-supervised features and training with small subsets of examples.

For the semi-supervised framework with a linear classifier ,(ii), we utilize the features of the labeled data and train a $d\times C$ linear classifier, where $d$ is the dimension of the features and $C$ is the number of classes.
To train the classifier, we use a learning rate of 2.5 and 400 epochs.
For the semi-supervised framework (iii), we train FlexMatch and SimMatch with a Wide-ResNet-28-10 model using SGD with momentum for 1000k iterations.
We employ a learning rate of 0.03, a batch size of 64, and a weight decay of $5\times 10^{-4}$.
Weak augmentations such as random crop and random horizontal flip are applied, while strong augmentations are obtained using RandAugment \cite{cubuk2020randaugment}.

\begin{table}[t]
\centering
\begin{adjustbox}{width=1\linewidth}
\begin{tabular}{lcccccc}
\toprule
Method & CIFAR-10 & CIFAR-100 & TinyImageNet & ImageNet & OrganMNIST & Yelp \\ \midrule
Random & 19.12 & 8.09 & 4.26 & 1.88 & 40.40 & 24.26 \\
Norm & 21.60 & 9.41 & 4.81 & 1.99 & 44.05 & 27.50 \\
Gram-Schmidt & 22.61 & 11.10 & 4.52 & - & 41.66 & 29.30 \\ \midrule
TypiClust & 24.45 & 10.78 & 5.61 & 4.94 & 46.58 & 24.26 \\
TypiClust\tiny{+norm} & 24.48 & 11.22 & 6.02 & 4.93 & 45.79 & \textbf{30.85} \\
ProbCover & 25.75 & 12.73 & \textbf{7.32} & 5.69 & 45.41 & - \\
ProbCover\tiny{+norm} & \textbf{27.43} & \textbf{13.10} & 7.29 & \textbf{5.76} & \textbf{47.99} & - \\ \bottomrule
\end{tabular}
\end{adjustbox}
\caption{Average accuracy of the 6 budgets used for the fully-supervised setting. See the average benefit of using norm.}
\label{tab:avg_fs}
\vspace{-0.5cm}
\end{table}

\begin{figure*}[t]
  \includegraphics[width=1\textwidth]{fs/legend_horizontal.pdf}\\
    \begin{subfigure}{0.24\linewidth}
    \includegraphics[width=1\textwidth]{fs/organ_fs.pdf}
    \caption{OrganMNIST}
    \label{fig:fs_organ}
  \end{subfigure}
  \begin{subfigure}{0.24\linewidth}
    \includegraphics[width=1\textwidth]{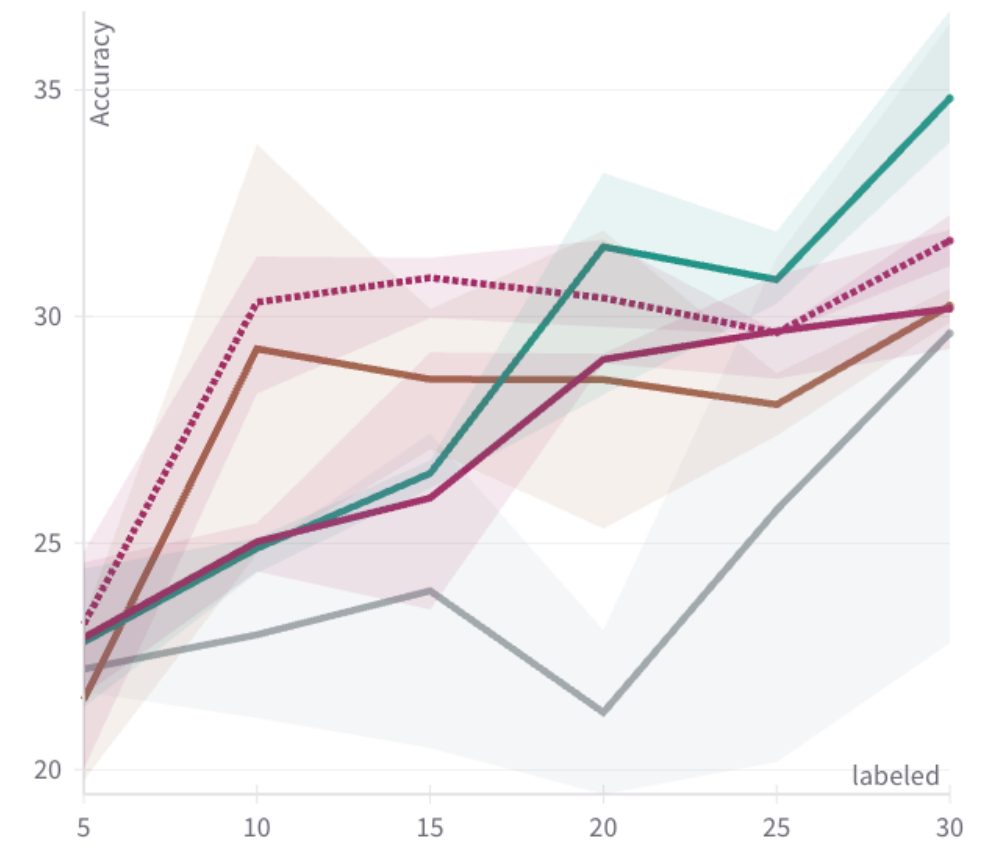}
    \caption{Yelp}
    \label{fig:yelp_fs}
  \end{subfigure}
  \begin{subfigure}{0.24\linewidth}
    \includegraphics[width=1\textwidth]{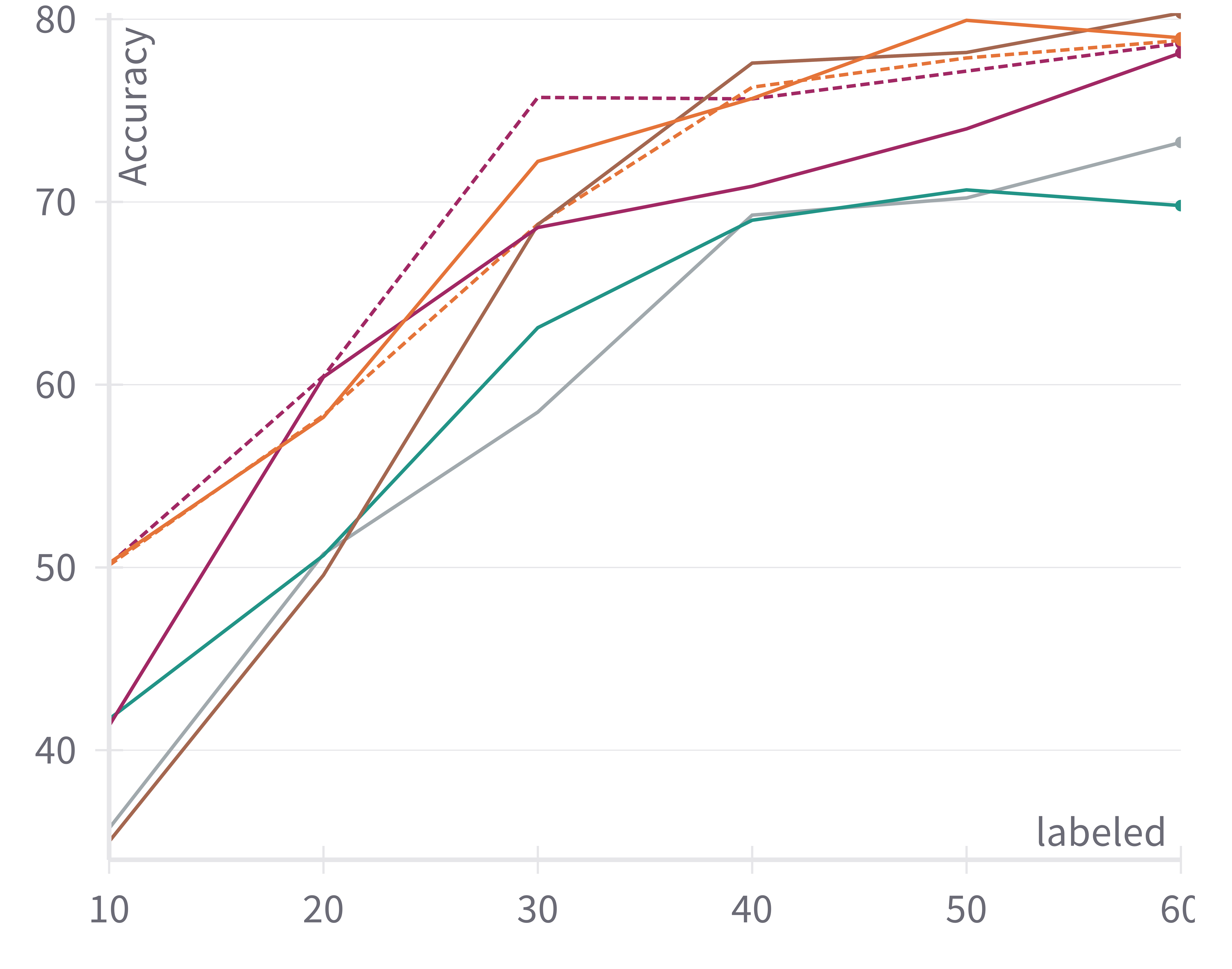}
    \caption{CIFAR-10}
    \label{fig:linear_cifar10}
  \end{subfigure}
  \begin{subfigure}{0.24\linewidth}
    \includegraphics[width=1\textwidth]{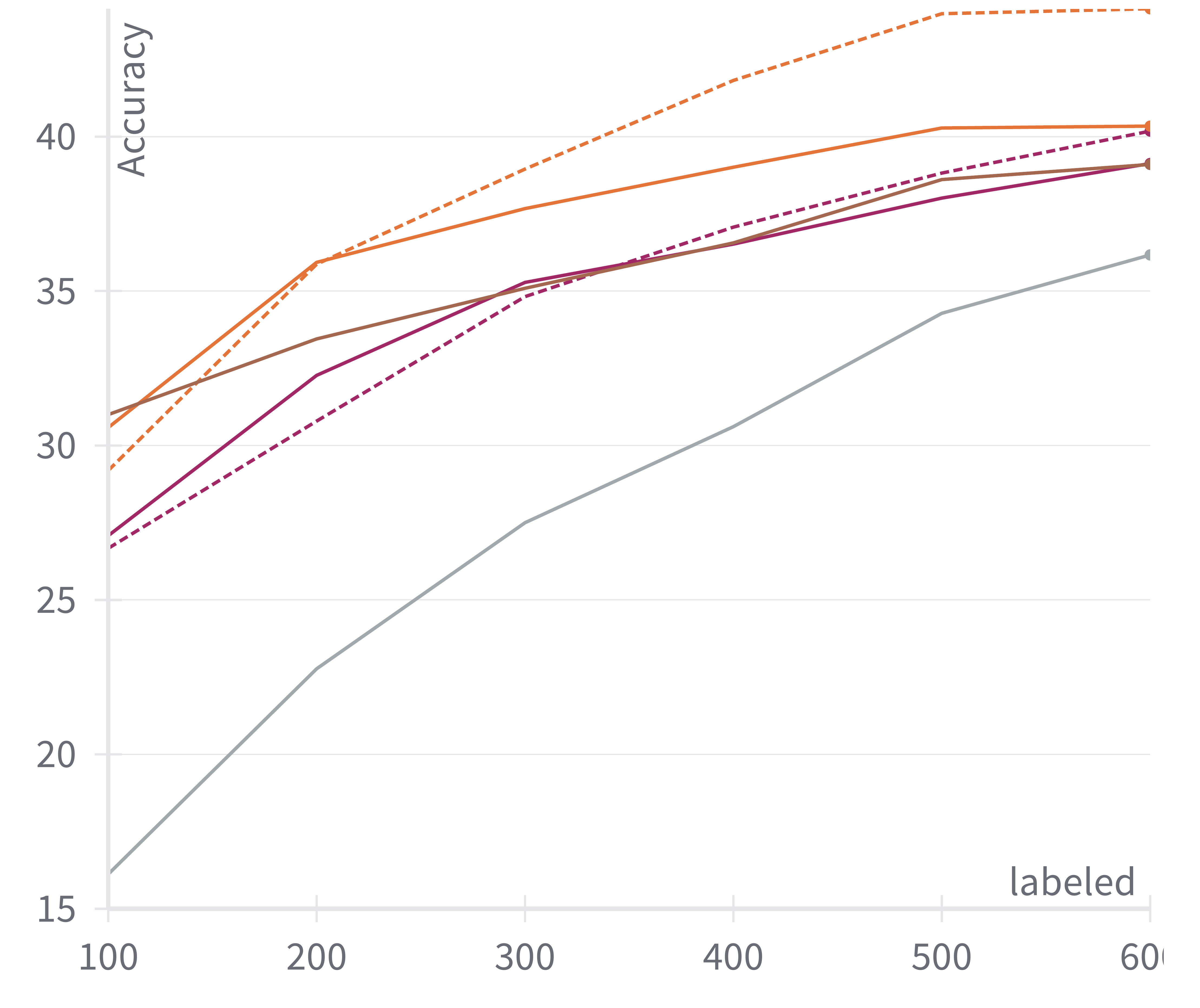}
    \caption{CIFAR-100}
    \label{fig:linear_cifar100}
  \end{subfigure} 
  \caption{\textbf{Fully-supervised framework:} Figs. (a) and (b) include results in the fully-supervised setting with OrganMNIST and fine-tuning of BERT with Yelp. Notice that also in this case, our approach provides benefits for subset selection. \textbf{Semi-supervised with linear classifier:} Figs. (c) and (d) include results with a linear classifier trained on top of self-supervised features. 
  }
  \label{fig:linear}\label{fig:add_datasets_fs_norm_dist}
\vspace{-0.6cm}
\end{figure*}

\subsection{Results}
\noindent\textbf{Fully supervised results. } The results in \cref{fig:fs} demonstrate that the simple baseline of randomization according to norm achieves a significant performance boost compared to uniform random selection in CIFAR-10, CIFAR-100 and Tiny-ImageNet.
The results also show that our proposed GS algorithm contributes to CIFAR-10/100.
It also contributes an additional enhancement in performance compared to randomization based on norm alone.
This is especially notable for CIFAR-100, where the results are comparable with TypiClust.
Integrating the norm criterion on top of previous methods yields comparable and even better results, which is particularly evident with CIFAR-10/100 with ProbCover and with Tiny-ImageNet with TypiClust.
For ImageNet, using the norm criterion leads to comparable results with the baselines.
The results for OrganMNIST are presented in \cref{fig:fs_organ}. In this case, using the norm criterion improved accuracy when used alone and when is combined with ProbCover. However, it did not improve the results with TypiClust.
Note that the Gram-Schmidt technique usually outperforms random choice but not other baselines tested.

For Yelp (\cref{fig:yelp_fs}), the results exhibit variance, likely due to the imbalanced nature of the chosen subset and the unbalanced test set. Selecting an extremely small subset may lead to low class coverage. Additionally, while the training set is balanced, the test set is highly imbalanced, which can amplify the effect of under-represented labels and result in significant fluctuations in test accuracy.
We see that on average, the norm and Gram-Schmidt surpass the performance of random choice and that using TypiClust with norm is on-par and usually better then without the norm addition.
In \cref{fig:fs_organ}, we present results with OrganAMNIST. The findings suggest that incorporating the norm criterion and Gram-Schmidt algorithm is advantageous particularly for higher budgets.
Overall, even though we enhance performance with extremely small labeled sets, there is still a significant drop in performance compared to the more than 90\% accuracy achieved with the full dataset. 
Thus, although we improve over the state-of-the-art, there is still room for improvement.

\cref{tab:avg_fs} summarizes the average accuracy results across all datasets in the fully-supervised setting, averaged over all subset sizes. The results demonstrate that incorporating the norm-based criterion consistently enhances performance.
In all tested cases, except for TinyImageNet, applying the norm criterion alongside baseline methods yields the highest average accuracy. Moreover, the performance degradation is minimal, with the largest difference being only 0.03. In contrast, the smallest observed improvement is 0.27 for CIFAR-100, with even larger gains achieved for other datasets, highlighting the substantial benefits of the norm-based criterion.

\noindent\textbf{Semi-supervised learning with a linear classifier. }  
\cref{fig:linear} displays the results obtained with the semi-supervised framework, where only one layer of the network is optimized.
The results indicate that norm randomization performs similarly to or slightly better than uniform randomization.
Our subset selection method shows improvements primarily over norm randomization, especially with higher budgets for CIFAR-10 and CIFAR-100, where the results are comparable with the TypiClust method.
Additionally, integrating the norm criterion with the baselines, ProbCover and TypiClust,  generally enhances their performance.
It should be noted that the results surpass those of the fully supervised setting, indicating that the features used are highly informative. 
The extracted features can significantly enhance performance, even when only a limited set of labels is available.

\begin{table}[t]
    \centering
    \begin{adjustbox}{width=1\linewidth}
    \begin{tabular}{lccccc}
    \toprule
         & Random & \begin{tabular}[c]{@{}c@{}}Features\\ Norm\end{tabular} & ProbCover & \begin{tabular}[c]{@{}c@{}}ProbCover\\ + Norm\end{tabular} & \begin{tabular}[c]{@{}c@{}}ProbCover\\ + GS\end{tabular}\\ \midrule
         FlexMatch  & 61.66 & 63.84 &  63.7 & 65.51  &  \textbf{79.9}\\ 
         SimMatch  & 38.03 & 39.68 & 54.68  & 57.08  & \textbf{76.02} \\ \midrule
         FlexMatch & 17.81 & 22.8 & 35.68 & 33.93 & \textbf{40.04}
        \\ 
          SimMatch   & 17.1  & 22.3 & 36.34 & 34.56 & \textbf{40.88}\\ 
         \bottomrule
    \end{tabular}
    \end{adjustbox}
        \caption{Semi-supervised training, FlexMatch \cite{zhang2021flexmatch} and SimMatch \cite{zheng2022simmatch}, with CIFAR-10 (top) and CIFAR-100 (bottom). Using feature norms is better than uniform random choice and adding GS on top of ProbCover enhances accuracy.}
    \label{tab:semi-supervised}
\end{table}

\noindent\textbf{Semi-Supervised with full fine-tuning. } The results for semi-supervised training, where all the network is fine-tuned and pseudo-labeling is employed, are provided in \cref{tab:semi-supervised}.
Our findings demonstrate a clear improvement in accuracy when using the norm criterion. 
Randomization according to the norm increases performance and achieves better accuracy than the sophisticated ProbCover method in some cases.
Additionally, it appears that incorporating the Gram-Schmidt algorithm on top of ProbCover further enhances performance in this setting.

\begin{figure*}[t]
  \centering
  \includegraphics[width=0.48\textwidth]{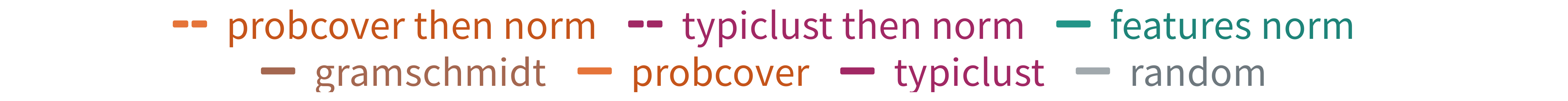} \includegraphics[width=0.48\textwidth]{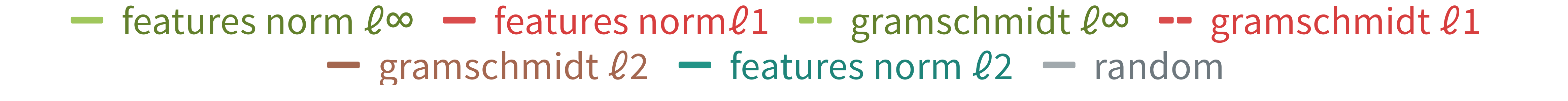} \\
  \begin{subfigure}{0.24\linewidth}
\includegraphics[width=1\textwidth]{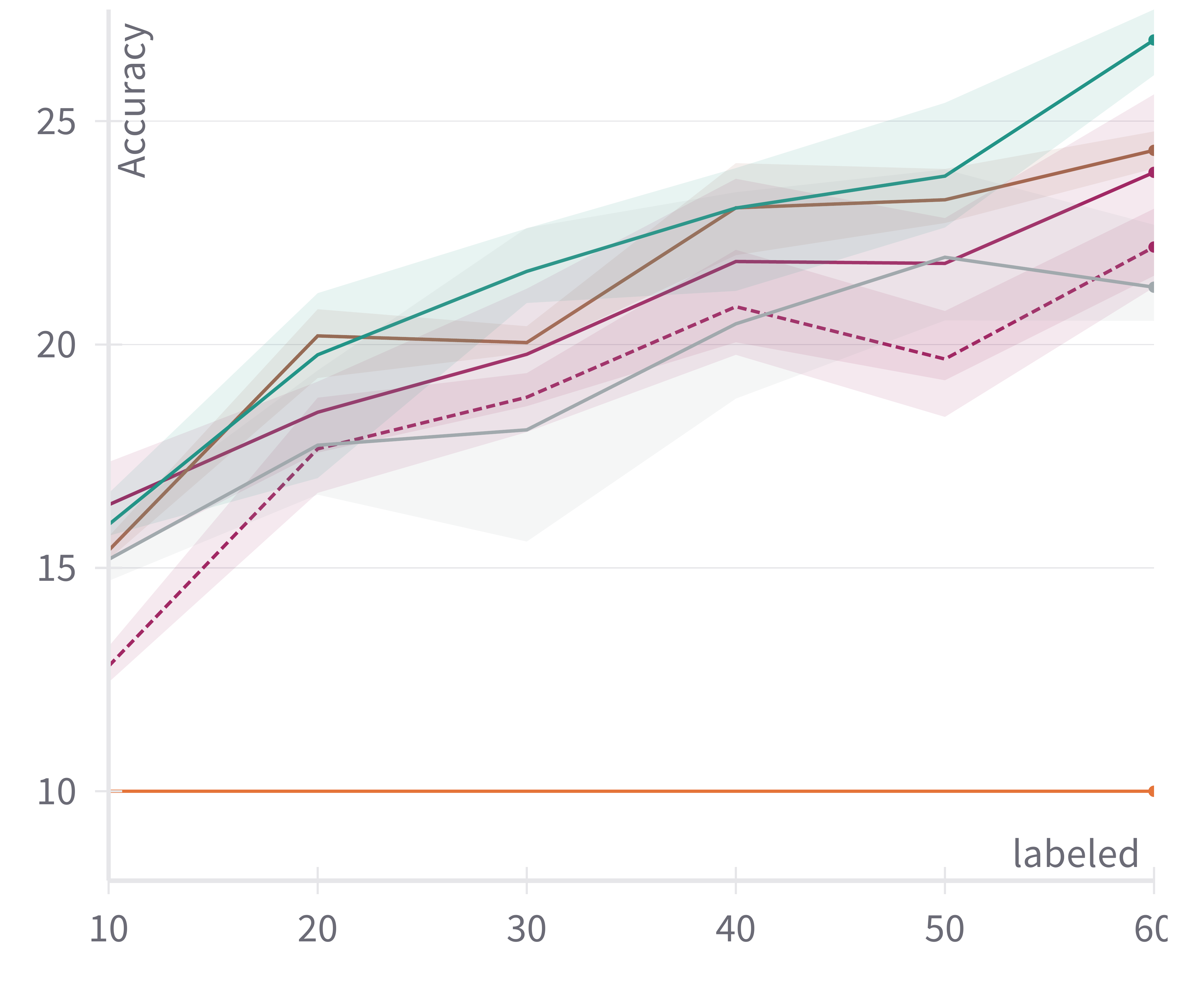}
    \caption{Randomly initialized network.}
    \label{fig:init_cifar10_fs}
  \end{subfigure}
  \begin{subfigure}{0.24\linewidth}
   \includegraphics[width=1\textwidth]{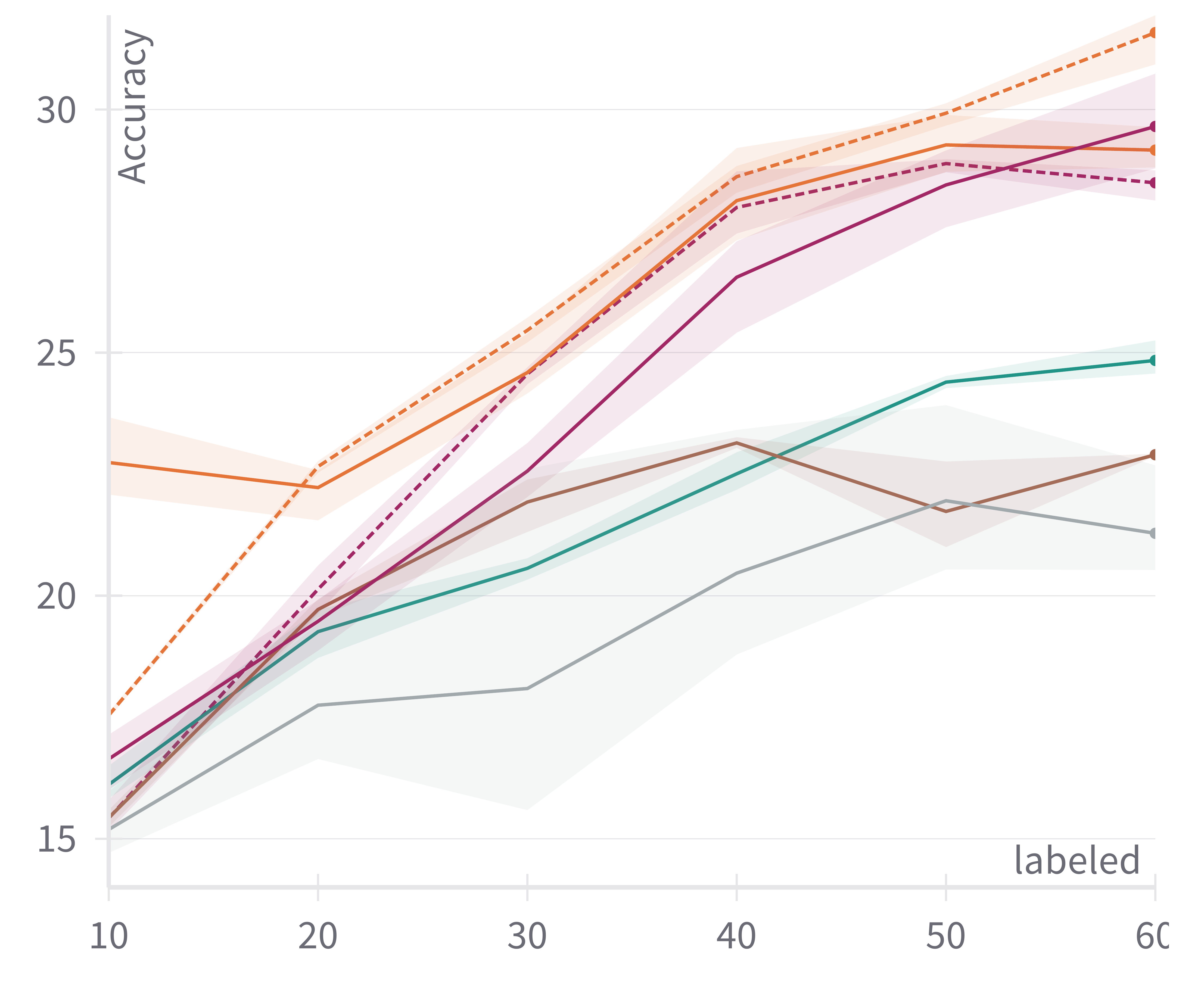}
    \caption{Self-supervised DINO.}
    \label{fig:dino_cifar10_fs}
  \end{subfigure} 
  \begin{subfigure}{0.24\linewidth}
    \includegraphics[width=1\textwidth]{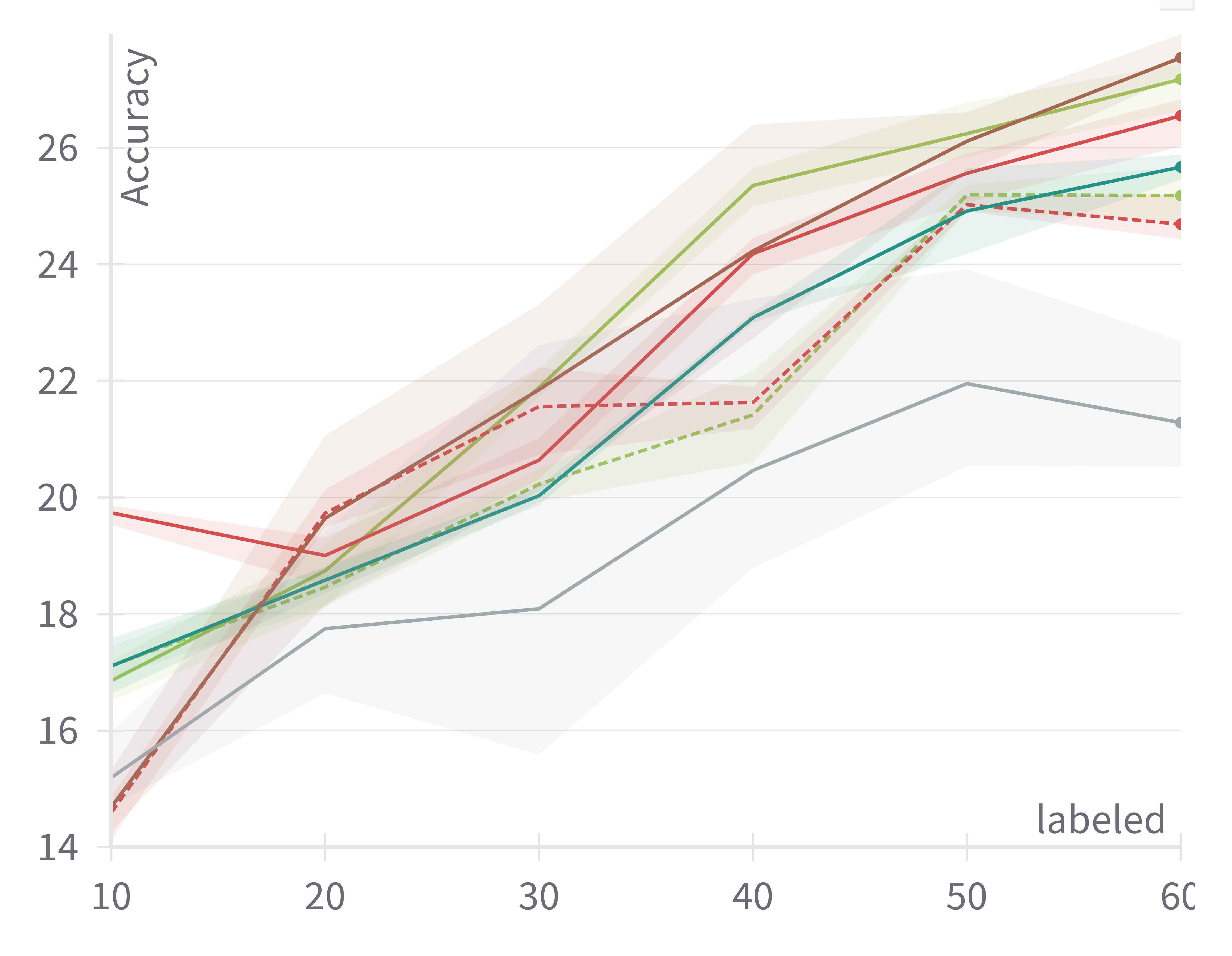}
    \caption{CIFAR10 with $\ell_1, \ell_\infty$ norms.}
    \label{fig:norms_cifar10_fs}
  \end{subfigure}
  \begin{subfigure}{0.24\linewidth}
    \includegraphics[width=1\textwidth]{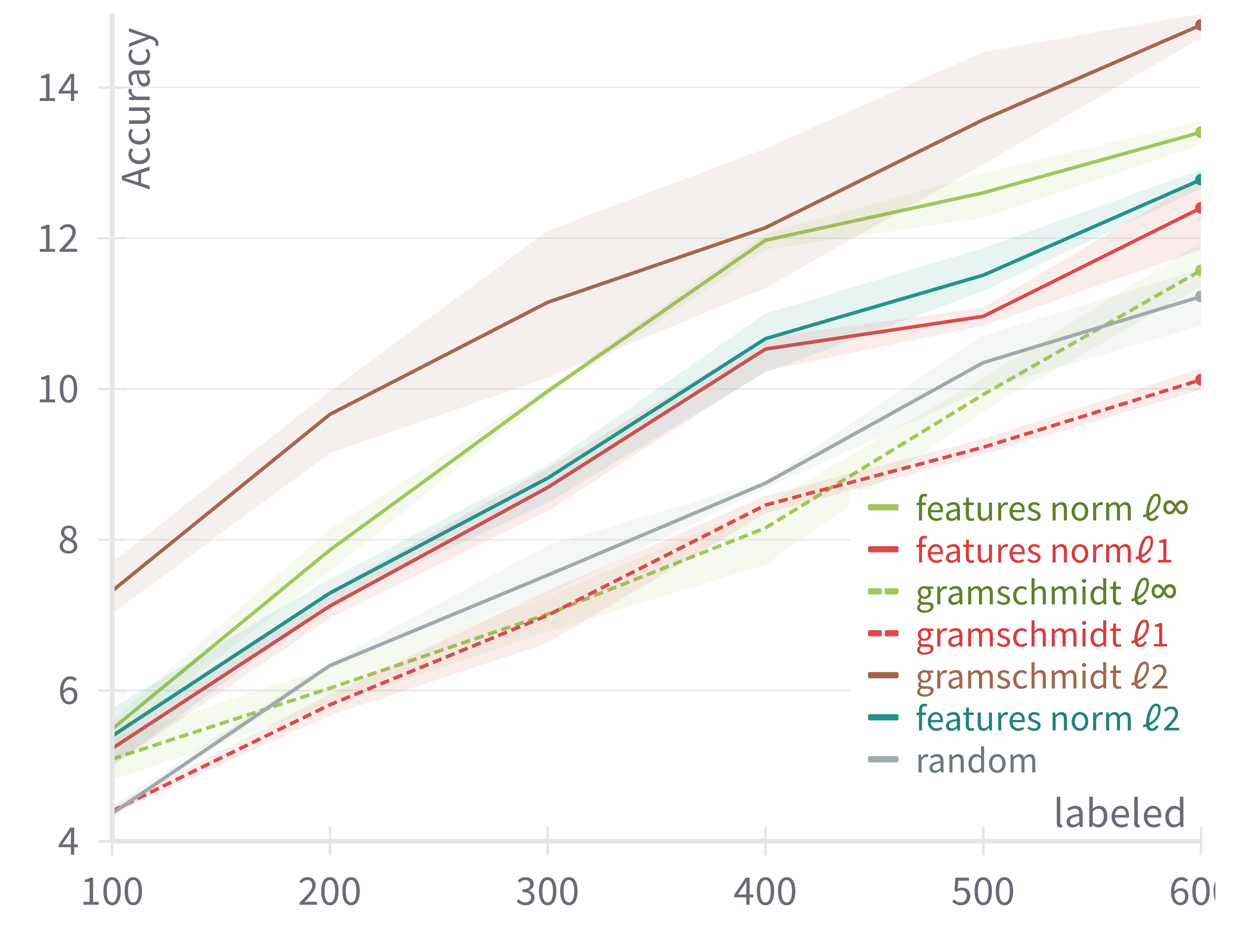}
    \caption{CIFAR100 with $\ell_1, \ell_\infty$ norms.}
    \label{fig:norms_cifar100_fs}
  \end{subfigure}
  \hfill
  \caption{\textbf{Feature domain.} \cref{fig:init_cifar10_fs,fig:dino_cifar10_fs} include the results with randomly initialized NN and DINO features. They show that
  the benefits of the feature norm apply to various feature domains. \textbf{Norms type.} 
  \cref{fig:norms_cifar10_fs} compares the results of using randomization based on feature norm and Gram-Schmidt algorithm, where the $\ell_1$ and $\ell_\infty$ are used instead of $\ell_2$ for the norm. Gram-Schmidt with $\ell_2$ provides the best results. For feature norm-based selection, $\ell_2$ and $\ell_\infty$ are the best. 
  The experimets are performed in the fully supervised setting.
  }
  \label{fig:ablation}
\end{figure*}

\subsection{Ablation Study}
\noindent\textbf{Dependency on the feature embedding. }
To ensure that the benefits of our method are not dependent on a specific feature domain, we conducted experiments with other embeddings.
Specifically, we employ a self-supervised approach, DINO \cite{caron2021emerging}, which is known for its informative features.
We experiment with the fully-supervised framework with CIFAR-10.
The results in \cref{fig:dino_cifar10_fs} indicate that utilizing norm randomization is beneficial.
Moreover, the addition of norm on top of existing methods mostly enhances the performance. 
Also, for the Gram-Schmidt method, we observe a performance gain, particularly at low budgets.

Given the computational demands and potential unavailability of pre-trained self-supervised models, we conducted experiments with features induced by randomly initialized  networks with CIFAR-10. 
The results with initialized neural networks are presented in \cref{fig:init_cifar10_fs}.
We observed that the performance of ProbCover collapsed and suffered from accuracy comparable to a random classifier.

The benefits of TypiClust over random sampling are limited in this case, and using the norm criterion did not improve this.
Importantly, that this is consistent with the claim made by Yehuda et al. \cite{yehuda2022active}, which clearly stated that TypiClust and ProbCover rely on informative features and suffer from poor performance with the RGB space. 
Our approach demonstrates benefits in both scenarios. Using the Gram-Schmidt algorithm is superior to random sampling and, notably, using the norm criterion alone induces high accuracy.
Although other methods experience a degradation in accuracy when using less informative features, we observe that randomization according to the norm maintains its performance.
Fig. 4 (in sup. mat.) includes results with the raw RGB space.

\noindent\textbf{Other norm types. }
In this work, we mainly rely on the $\ell_2$ norm but other types of norms may be considered.
Hence, we present results with $\ell_1$ and $\ell_\infty$ norms with CIFAR-10 and CIFAR-100 in \cref{fig:norms_cifar10_fs}.
We observe that randomization according to $\ell_\infty$ yields good results.
The performance of Gram-Schmidt with norms other than the $\ell_2$ norm harms performance, suggesting that our method suits the Euclidean norm.
Randomizing according to $\ell_1$ is better than $\ell_2$ for CIFAR-10 but slightly worse for CIFAR-100.
We employ the $\ell_2$ norm due to its consistent performance across scenarios.

\noindent\textbf{Necessity of randomization.} 
We investigate the benefits of randomization for successful subset selection. 
In Fig.6 (in supp. mat.), we present results, comparing scenarios where examples with the maximal norm are selected.
The results indicate that for selection based solely on norms, randomization is crucial for achieving accuracy gains, as accuracy only marginally surpasses that of a random classifier.

\noindent\textbf{Model agnostic selection. } 
Most of the results we present are based on SimCLR with ResNet backbone model as a feature extractor and training in the fully supervised setting, which is also performed with ResNet. To test the dependency on a specific architecture and demonstrate the model-agnostic nature of our method, we include results when the fully supervised training is performed with MobileNet \cite{howard2017mobilenets} with the selected subsets. The results appear in Fig. 7 (in supp. mat.). They further confirm that the norm criterion consistently enhances performance, when it used for training a different model. 

\noindent\textbf{Replace the norm criterion with random sampling. } 
To better assess the gain in performance for Typiclust and ProbCover that is achieved by the addition of our norm criterion, we replace the norm criterion with random sampling.
The results presented in Fig. 5 (in supp. mat.) indicate that the use of random sampling with the sets chosen by TypiClust and ProbCover leads to either degradation or comparable results. This is in contrast to the benefits of random selection based on the feature norm values.

\subsection{Discussion and Limitations} 
The results presented above highlight that employing a simple criterion such as the norm can yield non-trivial performance. Importantly, the norm criterion is easy to integrate into existing methods, usually enhancing or maintaining competitive performance with minimal additional computational overhead. Yet, in some minority cases, our approach leads to a small degradation in performance. Overall, on average, the embedding of norm criterion boosts performance.

The Gram-Schmidt approach also demonstrates improvement compared to uniformly random subset selection and norm-based selection. However, it falls short of the performance achieved by more sophisticated baselines, albeit with the advantage of lower computational complexity.

Through an extensive ablation study, we validated the robustness of our approach to variations in the norm order, feature domains, and model architecture. Our analysis confirms that the norm is indeed a beneficial measure for randomly chosen subsets and that the distribution of the norms carries meaningful information, rather than being vacuous.

\section{CONCLUSION AND FUTURE WORK}
We demonstrated a correlation between output feature norms and model accuracy, underscoring the importance of feature norms in the selection process. Moreover, we combined the norm criterion with the Gram-Schmidt process to improve the coverage of the feature space.

Extensive evaluations across diverse settings and datasets validate our approach’s efficacy. Compared to existing subset selection techniques, such as uniform random sampling, TypiClust, and ProbCover, our framework consistently improves performance, particularly when selecting extremely small subsets selected from large unlabeled pools. In many cases, our approach gets new state-of-the-art results, underscoring its practical utility in data-scarce scenarios like medical applications.

Looking ahead, this work opens several research avenues.
Promising direction involve incorporating our method into paradigms like transfer learning and semi-supervised learning in which  it may further enhance performance in resource-constrained environments.
Overall, we believe that our work advances the field of subset selection and lays a strong foundation for more effective and informative annotation strategies in deep learning.

\bibliographystyle{ieeetr}
\bibliography{sub_sel_short}

\clearpage
\appendix
\section{Additional Ablation Study}

\begin{figure}[h]
    \centering
    \includegraphics[width=0.8\linewidth]{fs/raw_rgb_legend.pdf}\\
    \includegraphics[width=0.8\linewidth]{fs/cifar10_raw_rgb.pdf}
    \caption{\textbf{RGB Domain.} Results with raw RGB values of the image used as features. The results  demonstrate advantages for both norm randomization and Gram-Schmidt algorithm compared with the baselines.} 
    \label{fig:raw-rgb}
\end{figure}

\begin{figure}[h]
  \centering
    \includegraphics[width=0.8\linewidth]{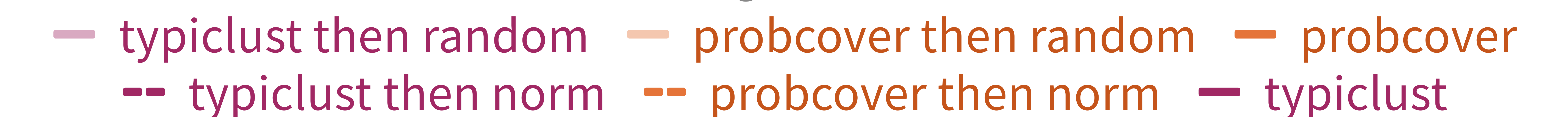}
    \includegraphics[width=0.8\linewidth]{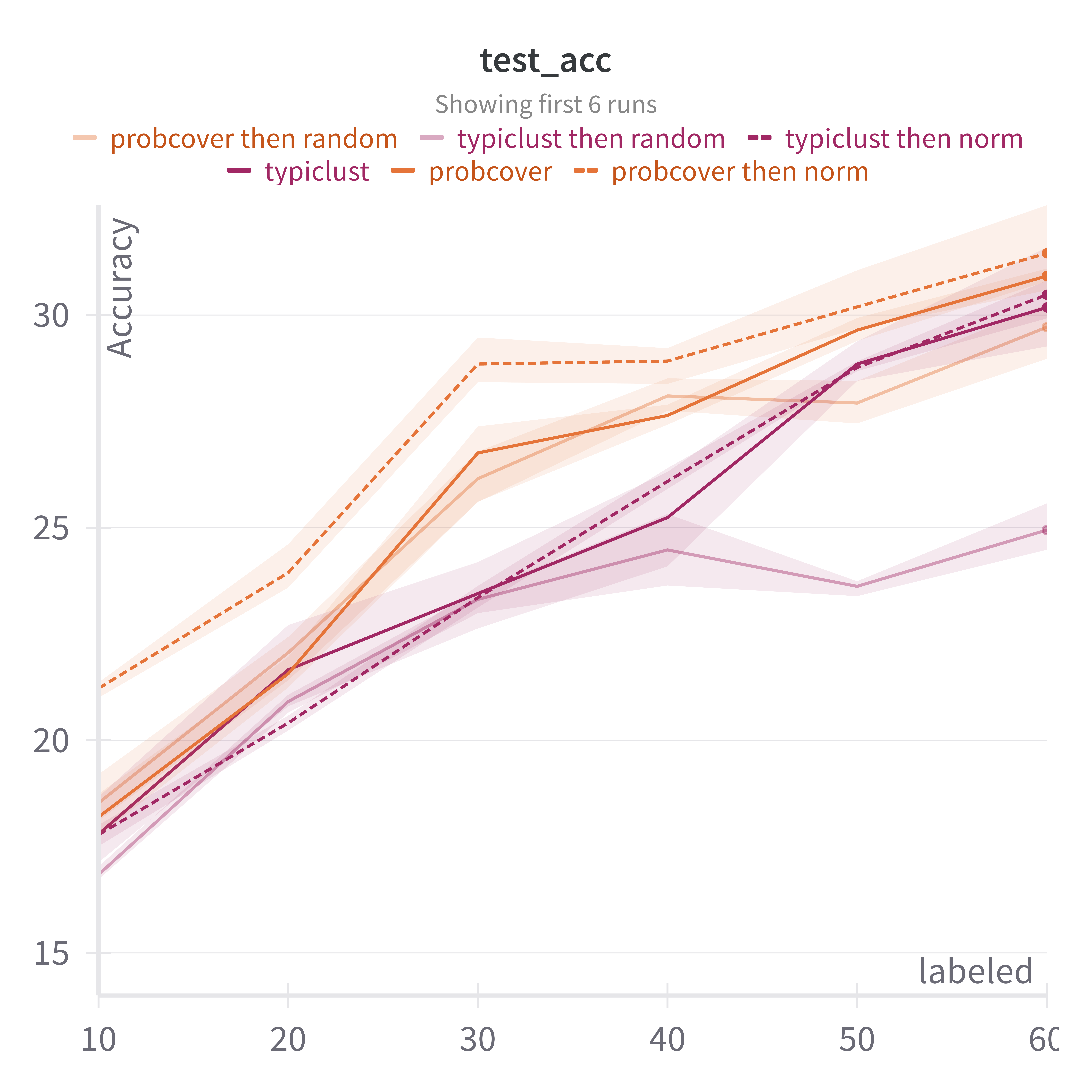}
  \caption{\textbf{Replace the Norm Criterion with Random Sampling. }The figure includes results obtained with uniform randomization rather than randomization according to the feature norm. First, a large set of examples is selected from CIFAR-10 using the baselines, and then uniform or norm-based classification is applied. The results show that randomization based on feature norms is pivotal for the enhancement in performance.}
  \label{fig:cifar10_then_random}
\end{figure}

  \begin{figure}
  \includegraphics[width=0.8\linewidth]{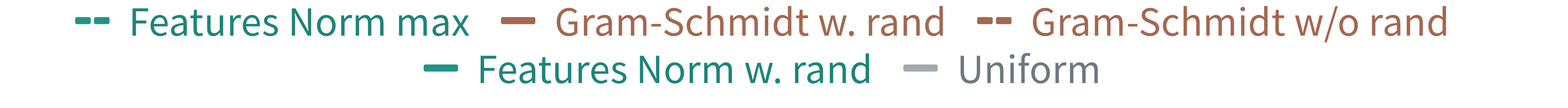}\\
    \includegraphics[width=0.8\linewidth]{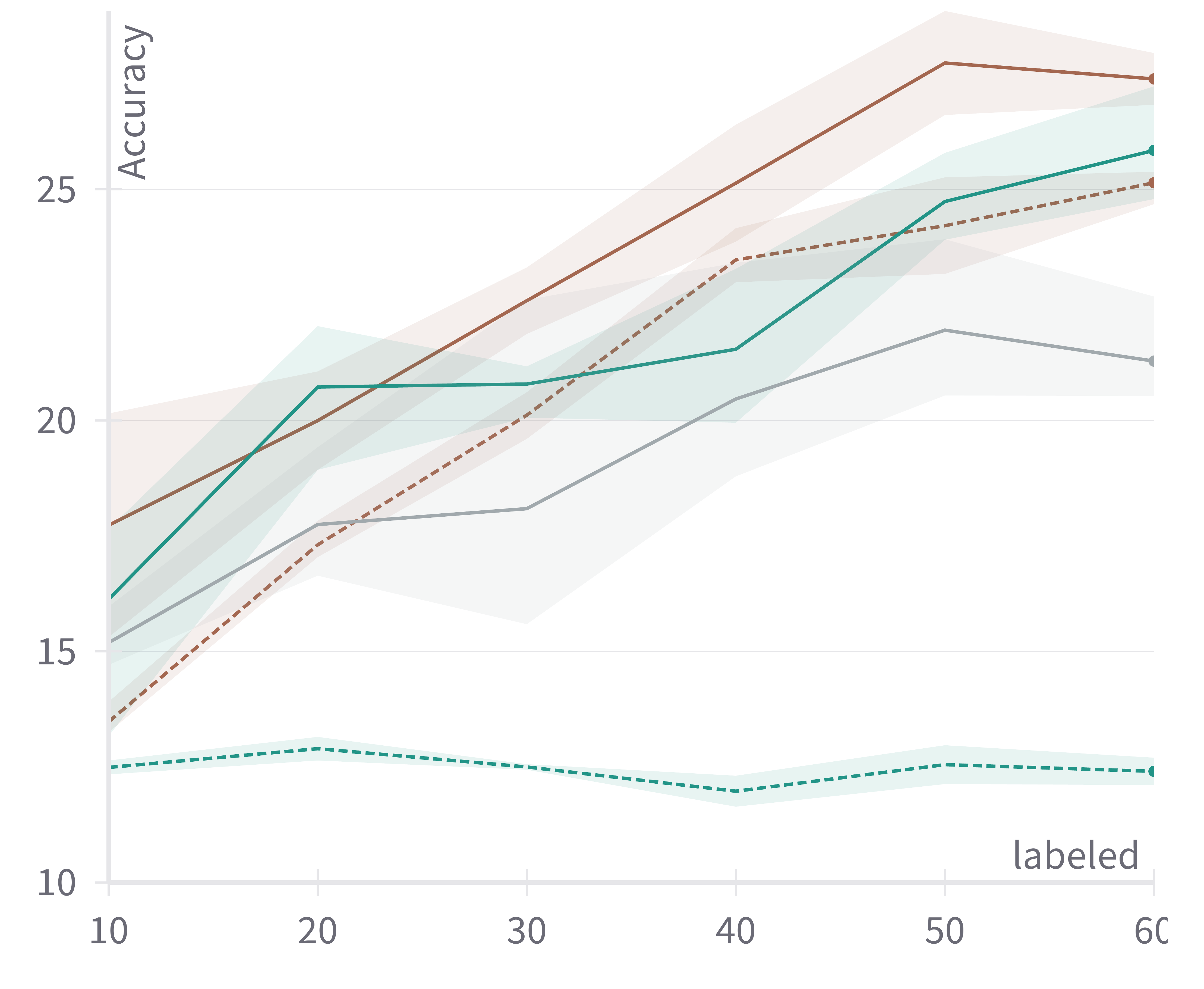}
\caption{\textbf{Necessity of Randomization.} We compare scenarios where examples with the maximal norm are chosen based on feature norms and with our Gram-Schmidt algorithm.
Specifically, in our algorithm (\cref{algorithm}), the randomization step is replaced with selecting the example with the highest feature norm.
The results clearly indicated that randomization is crucial for performance since selecting the maximal norms leads to performance that is comparable to that of a random classifier.}\label{fig:randomization_norm_gs}
  \end{figure}

\begin{figure}[h]
    \centering    \includegraphics[width=0.8\linewidth]{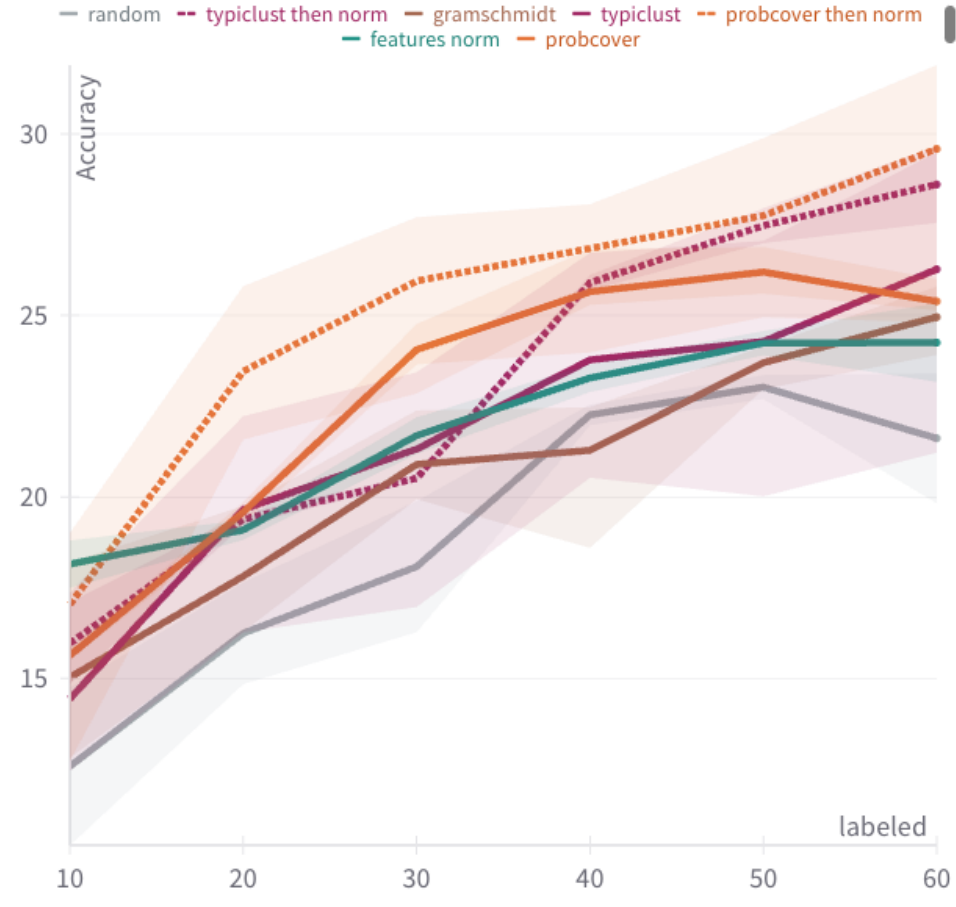}
    \caption{\textbf{Model Agnostic Selection.} We include results with fully-supervised setting with MobileNet \cite{howard2017mobilenets} and CIFAR-10.
The results present that our model is using norm criterion indeed boosts performance. For Gram-Schmidt the performance is better or on-par with random subsets. }
    \label{fig:mobilenet}
\end{figure}

\end{document}